\documentclass{article} 
\usepackage{iclr2015,times}
\usepackage{hyperref}
\usepackage{url}

\title{Ensemble of Generative and Discriminative Techniques for Sentiment Analysis of Movie Reviews}

\author{
    Gr\'{e}goire Mesnil \\
    University of Montr\'{e}al \\
    University of Rouen
\And
Tomas Mikolov \& Marc'Aurelio Ranzato \\ 
Facebook Artificial Intelligence Research
\And
Yoshua Bengio \\
University of Montr\'{e}al
}

\iclrfinalcopy 

\begin{document}

\maketitle

\begin{abstract} Sentiment analysis is a common task in natural language
processing that aims to detect polarity of a text document (typically a
consumer review). In the simplest settings, we discriminate only between
positive and negative sentiment, turning the task into a standard binary
classification problem. We compare several machine learning approaches to this
problem, and combine them to achieve a new state of the art. We show how to
use for this task the standard generative language models, which are slightly
complementary to the state of the art techniques. We achieve strong results on
a well-known dataset of IMDB movie reviews. Our results are easily
reproducible, as we publish also the code needed to repeat the experiments.
This should simplify further advance of the state of the art, as other
researchers can combine their techniques with ours with little effort.
\end{abstract}

\section{Introduction}

Sentiment analysis is among the most popular, simple and useful tasks in
natural language processing. It aims at predicting the attitude of text,
typically a sentence or a review. For instance, movies or restaurant are often
rated with a certain number of stars, which indicate the degree to which the
reviewer was satisfied.

This task is often considered as one of the simplest in NLP because basic
machine learning techniques can yield strong baselines \citep{Wang2012},
often beating much more intricate approaches \citep{Socher2011}. In the simplest
settings, this task can be seen as a binary classification between positive and
negative sentiment.

However, there are several challenges towards achieving the best possible
accuracy. It is not obvious how to represent variable length documents beyond
simple bag of words approaches that lose word order information. One can use
advanced machine learning techniques such as recurrent neural networks and
their variations \citep{Mikolov2010, Socher2011}, however it is not clear if
these provide any significant gain over simple bag-of-words and bag-of-ngram
techniques \citep{Pang2008, Wang2012}.

In this work, we compared several different approaches and realized, without
much surprise, that model combination performs better than any individual
technique. The ensemble best benefits from models that are complementary, thus
having diverse set of techniques is desirable. The vast majority of models
proposed in the literature are discriminative in nature, as their parameters
are tuned for the classification task directly. In this work, we boost the
performance of the ensemble by considering a generative language model. To this
end, we train two language models, one on the positive reviews and one on the
negative ones, and use the likelihood ratio of these two models evaluated on
the test data as an additional feature. For example, we assume that a positive
review will have higher likelihood to be generated by a model that was trained
on a large set of positive reviews, and lower likelihood given the negative
model. In this paper, we constrained our work to binary classification where we
trained two generative models, positive and negative. One could consider a
higher number of classes since this approach scales linearily with the number
of models to be train, i.e. one for each class.   The large pool of diverse
models is a) simple to implement (in line with previous work by Wang and
Manning \citep{Wang2012}) and b) it yields state of the art performance on
one of the largest publicly available benchmarks of movie reviews, the Stanford
IMDB dataset of reviews. Code to reproduce our experiments is available at
\url{https://github.com/mesnilgr/iclr15}.

\section{Description of the Models}

In this section we describe in detail the approaches we considered in our
study. The novelty of this paper consists in combining both generative and
discriminative models together for sentiment prediciton. 

\subsection{Generative Model}

A generative model defines a distribution over the input. By training a
generative model for each class, we can then use Bayes rule to predict which
class a test sample belongs to. More formally, given a dataset of pairs $\{x^{(i)},
y^{(i)}\}_{i=1,\dots, N}$ where $x^{(i)}$ is the $i$-th document in the training set, $y^{(i)} \in
\{-1,+1\}$ is the corresponding label and N is the number of training samples, we
train two models: $p^{+}(x | y = +1)$ for $\{x^{(i)} \textrm{~subject to~} y^{(i)} = +1\}$ and $p^{-}(x | y = -1)$ for
$\{x \textrm{~subject to~} y = -1\}$. Then, given an input $x$ at test time we compute the ratio
(derived from Bayes rule): $r =  p^{+}(x | y = +1) / p^{-}(x | y = -1) \times p(y = +1) /
p(y = -1)$. If $r > 1$, then $x$ is assigned to the positive class, otherwise to the
negative class.

We have a few different choices of distribution we can choose from. The most
common one is the n-gram, a count-based non-parametric method to compute
$p(x^{(i)}_{k} | x^{(i)}_{k-1}, x^{(i)}_{k-2}, \dots, x^{(i)}_{k-N+1})$, where $x^{(i)}_{k}$ is the $k$-th word
in the $i$-th document. In order to compute the likelihood of a document, we use
the Markov assumption and simply multiply the n-gram probabilities over all
words in the document: $p(x^{(i)}) = \prod_{k=1}^{K} p(x^{(i)}_{k} | x^{(i)}_{k-1}, x^{(i)}_{k-2},
\dots , x^{(i)}_{k-N+1})$. As mentioned before, we train one n-gram language model
using the positive documents and one model using the negative ones.

In our experiments, we used SRILM toolkit \citep{Srilm2002} to train the n-gram language
models using modified Kneser-Ney smoothing \citep{Kneser1995}. Furthermore, as both language
models are trained on different datasets, there is a mismatch between
vocabularies: some words can appear only in one of the training sets. This can
be a problem during scoring, as the test data contain novel words that were not
seen in at least one of the training datasets. To avoid this problem, it is
needed to add penalty during scoring for each out of vocabulary word.

N-grams are a very simple data-driven way to build language models. However,
they suffer from both data sparsity and large memory requirement. Since the
number of word combinations grows exponentially with the length of the context,
there is always little data to accurately estimate probabilities for higher
order n-grams.

In contrast with N-grams languages models, Recurrent neural networks
(RNNs)~\citep{Mikolov2010} are parametric models that can address these issues. The
inner architecture of the RNNs gives them potentially infinite context window,
allowing them to perform smoother predictions. We know that in practice, the
context window is limited due to exploding and vanishing gradients \citep{Pascanu2012}.
Still, RNNs outperform significantly n-grams and are the state of the art for
statistical language modeling. A review of these techniques is beyond the scope
of this short paper and we point the reader to~\citep{MikolovPhd}
for a more in depth discussion on this topic.

Both when using n-grams and RNNs, we compute the probability of the test
document belonging to the positive and negative class via Bayes' rule. These
scores are then averaged in the ensemble with other models, as explained in
Section~\ref{sec:ensemble}.

\subsection{Linear classification of weighted n-gram features}

Among purely discriminative methods, the most popular choice is a linear
classifier on top of a bag-of-word representation of the document. The input
representation is usually a tf-idf weighted word counts of the document. In
order to preserve local ordering of the words, a better representation would
consider also the position-independent n-gram counts of the document
(bag-of-n-grams).

In our ensemble, we used a supervised reweighing of the counts as in the Naive
Bayes Support Vector Machine (NB-SVM) approach \citep{Wang2012}.  This approach computes
a log-ratio vector between the average word counts extracted from positive
documents and the average word counts extracted from negative documents.  The
input to the logistic regression classifier corresponds to the log-ratio vector
multiplied by the binary pattern for each word in the document vector. Note
that the logictic regression can be replaced by a linear SVM. Our
implementation\footnote{https://github.com/mesnilgr/nbsvm} slightly improved
the performance reported in \citep{Wang2012} by adding tri-grams (improvement
of $+0.6\%$), as shown in Table~\ref{tab:svm}.

\begin{table}[t]
\caption{Performance of SVM with \cite{Wang2012} rescaling for different N-grams}
\label{tab:svm}
\begin{center}
\begin{tabular}{ll}
\hline
Input features & Accuracy \\
\hline
Unigrams & $88.61\%$ \\
Unigrams+Bigrams & $91.56\%$ \\
Unigrams+Bigrams+Trigrams & $91.87\%$ \\
\hline
\end{tabular}
\end{center}
\end{table}

\subsection{Sentence Vectors}

Recently, \citep{Le2014} proposed an unsupervised method to
learn distributed representations of words and paragraphs. The key idea is to
learn a compact representation of a word or paragraph by predicting nearby
words in a fixed context window. This captures co-occurence statistics and it
learns embeddings of words and paragraphs that capture rich semantics. Synonym
words and similar paragraphs often are surrounded by similar context, and
therefore, they will be mapped into nearby feature vectors (and vice versa). 

Such embeddings can then be used to represent a new document (for instance, by
averaging the representations of the paragraphs that constitute the document)
via a fixed size feature vector.  The authors then use such a document
descriptor as input to a one hidden layer neural network for sentiment
discrimination.

\subsection{ Model Ensemble \label{sec:ensemble}}

In this work, we combine the log probability scores of the above mentioned
models via linear interpolation. More formally, we define the overall
probability score as the weighted geometric mean of baseline models: $p(y=+1 |
x) = \prod p^{k}(y=+1 | x)^{\alpha_k}$, with $\alpha_k > 0$.

We find the best setting of weights via brute force grid search, quantizing the
coefficient values in the interval $[ 0, 1 ]$ at increments of $0.1$. The search is
evaluated on a validation set to avoid overfitting. We do not focus on a
smarter way to find the $\alpha$ since we consider only $3$ models in our approach
and we consider it out of the scope of this paper. Using more models would make
the use of such method prohibitive. For a larger number of models, one might
want to consider random search of the $\alpha$ coefficients or even Bayesian
approaches as these techniques will give better running time performance. 

\section{Results}

In this section we report results on one of the largest publicly available
sentiment analysis datasets, the IMDB dataset of movie reviews. The dataset
consists of $50,000$ movie reviews which are categorized as being either positive
or negative. We use $25,000$ reviews for training and the rest for testing, using
the same protocol proposed by \citep{Maas2011}.  All experiments can
be reproduced using the code available at \url{https://github.com/mesnilgr/iclr15}.


Table~\ref{tab:individual} reports the results of each individual model. We
have found that generative models performed the worst, with RNNs slightly
better than n-grams.  The most competitive method is the method based on
reweighed bag-of-words~\citep{Wang2012} \footnote{In our experiments, to match
the results from~\citep{Le2014}, we followed the suggestion by Quoc Le to use
hierarchical softmax instead of negative sampling.  However, this produces the
$92.6\%$ accuracy result only when the training and test data are not shuffled.
Thus, we consider this result to be invalid.}. Favoring simplicity and
reproducibility of our performance, all results reported in this paper were
produced by a linear classifier. 

\begin{table}[t]
\caption{Performance of Individual Models}
\label{tab:individual}
\begin{center}
\begin{tabular}{ll}
\hline
Single Methods & Accuracy \\
\hline
N-gram & $86.5\%$ \\
RNN-LM & $86.6\%$ \\
Sentence Vectors & $88.73\%$ \\
NB-SVM Trigram & $91.87\%$ \\
\hline
\end{tabular}
\end{center}
\end{table}

Finally, Table~\ref{tab:ensemble} reports the results of combining the previous
models into an ensemble. When we interpolate the scores of RNN, sentence
vectors and NB-SVM, we achieve a new state-of-the-art performance of $92.57\%$,
to be compared to $91.22\%$ reported by \citep{Wang2012}. Notice that our
implementation of the Sentence Vectors method \citep{Le2014} alone yielded only
$88.73\%$ (a difference of $\simeq 4\%$).  In order to measure the contribution
of each model to the final ensemble classifier, we remove one model at a time
from the ensemble. We observe that the removal of the generative model affects
the least the ensemble performance.  Overall, all three models contribute to
the success of the overall ensemble, suggesting that these three models pick up
complimentary features useful for discrimination. In Table~\ref{tab:sent}, we
show test reviews misclassified by single models but classified accurately by
the ensemble.

\begin{table}[t]
\caption{Performance of Different Model Combinations}
\label{tab:ensemble}
\begin{center}
\begin{tabular}{ll}
\hline
Ensemble & Accuracy \\
\hline
RNN-LM + NB SVM Trigram & $92.13\%$ \\ 
RNN-LM + Sentence Vectors & $90.4\%$ \\
Sentence Vectors + NB-SVM Trigrams & $92.39\%$ \\
{\bf All} & $\mathbf{92.57\%}$ \\
State of the art & $91.22\%$ \\
\hline
\end{tabular}
\end{center}
\end{table}

\begin{table}[t]
\caption{Reviews misclassified by Single Models but classified accurately by the Ensemble}
\label{tab:sent}
\begin{center}
\begin{tabular}{ll}
\hline
Model & Sentences \\
\hline
& {\bf (positive)} a really realistic ,  sensible movie by ramgopal verma  .  no stupidity like \\
NB-SVM        & songs as in other hindi movies .  class acting by nana patekar .    much similarities to \\
        & real 'encounters' . \\
        & {\bf  (negative)} leslie nielson is a very talented actor ,  who made a huge mistake by doing \\
        & this film .  it doesn't even come close to being funny .  the best word to describe \\
        & it is  stupid ! \\
\hline
&  {\bf (positive)} this is a good film .  this is very funny .  yet after this film there \\
RNN-LM        & were no good ernest films ! \\
        & {\bf  (negative)} a real hoot ,  unintentionally .  sidney portier's character \\
        & is so sweet and lovable you want to smack him .  nothing about this movie rings true .\\
        &  and it's boring to boot . \\
\hline
& {\bf   (positive)} this movie is based on the novel island of dr .  moreau by \\
Sentence Vector
                    & version by john frankenheimer . \\
                    & {\bf  (negative)} if it wasn't for the terrific music ,  i would not hesitate to give this  \\
                    & cinematic underachievement 2/10 .  but the music actually makes me like certain \\ 
                    &passages ,  and so i give it 5/10 . \\
\hline
\end{tabular}
\end{center}
\end{table}

\section{Conclusion}

We have proposed a very simple yet powerful ensemble system for sentiment
analysis. We combine three rather complementary and conceptually different
baseline models: one based on a generative approach (language models), one
based on continuous representations of sentences and one based on a clever
reweighing of tf-idf bag-of-word representation of the document. Each such
model contributes to the success of the overall system, achieving the new state
of the art performance on the challenging IMDB movie review dataset. Code to
reproduce our experiments is available at: \url{https://github.com/mesnilgr/iclr15}.
We hope researchers will take advantage of our code to include their new
results into our ensemble and focus on improving the state of the art for
Sentiment Analysis.




\bibliography{iclr2015}
\bibliographystyle{iclr2015}

\end{document}